\documentclass[letterpaper, 10 pt, conference]{ieeeconf}  

\IEEEoverridecommandlockouts                              

\usepackage{amsmath} 
\usepackage{amssymb}  

\usepackage{cite}
\usepackage[pdftex]{graphicx}
\usepackage{algorithm}
\usepackage{algorithmicx}
\usepackage{algpseudocode}
\algrenewcommand\algorithmicrequire{\textbf{Input:}}
\algrenewcommand\algorithmicensure{\textbf{Output:}}
\usepackage{array}
\usepackage{multirow}
\usepackage[font=small]{caption}
\usepackage{subcaption}

\usepackage{tikz}
\usepackage{pgfplots}
\usepackage{xcolor}
\pgfplotsset{compat=1.17}
\usetikzlibrary{positioning}
\usetikzlibrary{decorations.pathreplacing}
\usetikzlibrary{calc}
\usetikzlibrary{positioning,shapes,shadows,arrows}

\pgfplotsset{
every axis/.append style={
  axis line style={->}, 
  legend style={font=\scriptsize},
  label style={font=\scriptsize},
  title style={font=\scriptsize},
  tick label style={font=\scriptsize},
  }
}

\makeatletter
\let\NAT@parse\undefined
\makeatother
\usepackage{hyperref}

\title{\LARGE \bf
Kinodynamic Rapidly-exploring Random Forest for Rearrangement-Based Nonprehensile Manipulation
}

\author{Kejia Ren, Podshara Chanrungmaneekul, Lydia E. Kavraki and Kaiyu Hang
\thanks{The authors are with the Department of Computer Science, Rice University, Houston, TX 77005, USA. \{\tt\small kr43, pc45, kavraki, kaiyu.hang\}@rice.edu.}
\thanks{In this work, KR, PC and KH are supported by NSF CMMI-2133110 and Rice University Funds, and LEK is supported in part by NSF 2008720.}%
}

\begin{document}

\maketitle
\thispagestyle{empty}
\pagestyle{empty}

\begin{abstract}
Rearrangement-based nonprehensile manipulation still remains as a challenging problem due to the high-dimensional problem space and the complex physical uncertainties it entails. We formulate this class of problems as a coupled problem of local rearrangement and global action optimization by incorporating free-space transit motions between constrained rearranging actions.
We propose a forest-based kinodynamic planning framework to concurrently search in multiple problem regions, so as to enable global exploration of the most task-relevant subspaces, while facilitating effective switches between local rearranging actions. By interleaving dynamic horizon planning and action execution, our framework can adaptively handle real-world uncertainties. With extensive experiments, we show that our framework significantly improves the planning efficiency and manipulation effectiveness while being robust against various uncertainties.

\end{abstract}
\section{Introduction}
\label{sec:intro}


Many manipulation tasks require the robot to concurrently rearrange multiple objects through nonprehensile actions to achieve the goal, such as grasping in clutter, sorting, and singulation \cite{agboh2018real, huang2019large, moll2017randomized}.
%
%
%
%
%
Despite its importance, rearrangement-based manipulation still remains challenging. First, the problem is often computationally very expensive due to the high dimensional problem spaces and the involved complex physics computation. Second, there are always certain discrepancies between the physics models and the real world, resulting in inevitable execution failures after a manipulation plan is successfully found.

\begin{figure}[t]
    \centering
        \includegraphics[width=0.8\columnwidth]{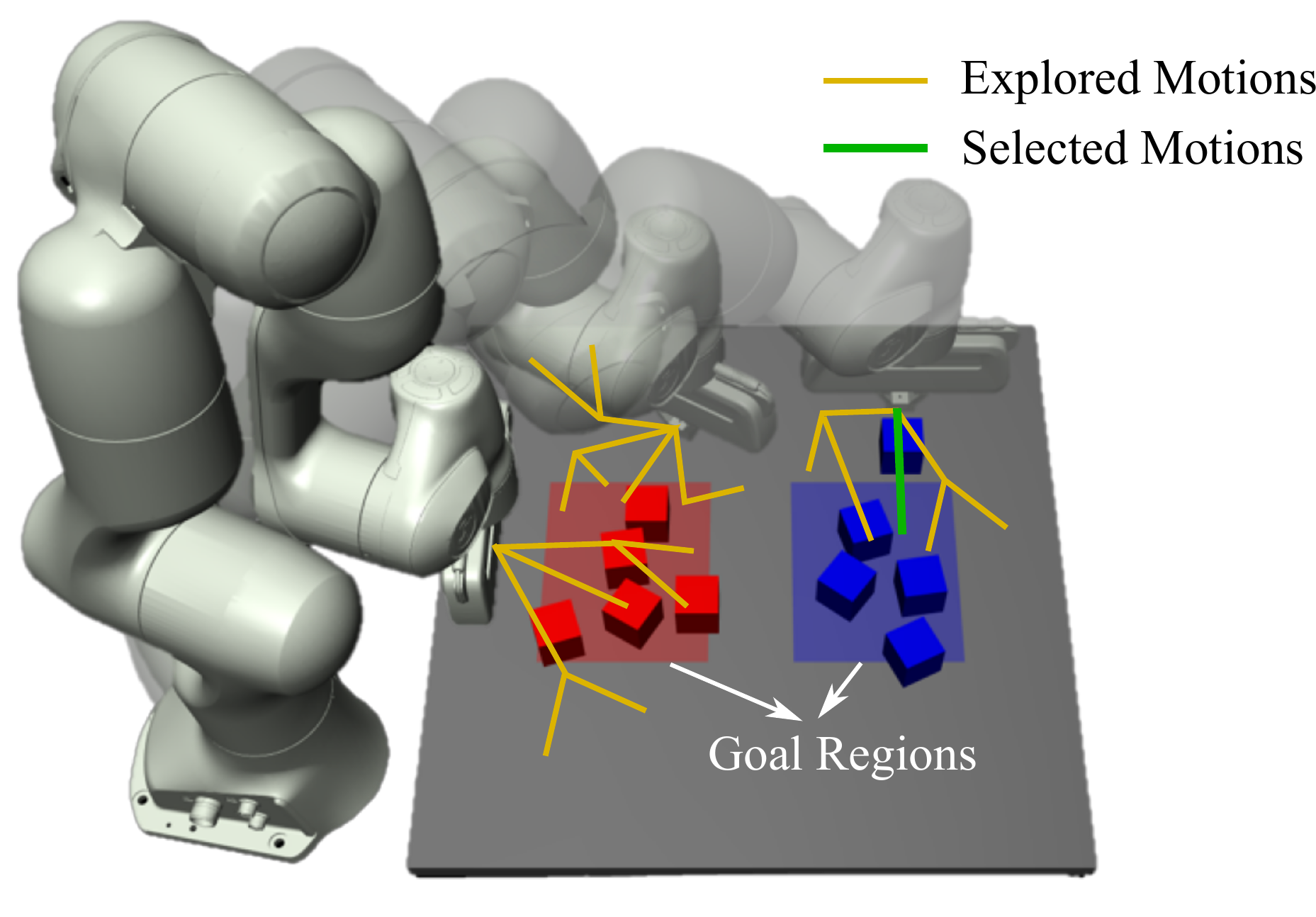}
    \caption{A motivational scenario where the robot is tasked to reposition each object to its corresponding rectangular goal region. By our forest-based framework, the robot is able to concurrently explore local motions (yellow) rooted at different configurations, and select the most efficient one (green) from a global perspective.}
    \label{fig:motivation}
    \vspace{-0.5cm}
\end{figure}

In our previous work, we proposed a planning framework termed as dhRRT\cite{ren2022rearrangement}. Although it has improved the planning efficiency and execution robustness, like many existing rearrangement planners\cite{king2015nonprehensile, haustein2015kinodynamic, king2017unobservable}, 
dhRRT simplifies the problem by restricting the robot motions to be always inside a task manifold such that the end-effector moves only within $SE(2)$.
However, such motion constraints can make the manipulation plans less efficient
as it significantly removes many possible actions by shrinking the solution space.
For the tabletop manipulation shown in Fig.~\ref{fig:motivation} as an example, the robot manipulator is tasked to reposition each object in different colors
to their corresponding rectangular goal regions.
The current robot configuration is shown by the opaque robot in the figure.
With the aforementioned motion constraints,
the robot will explore very locally, therefore, will be unable to quickly switch to a more task-effective configuration and explore from there to find more efficient motions, such as the the pushing action illustrated by the green line in Fig.~\ref{fig:motivation}.

In this work, for enabling the robot to explore multiple problem subspaces and optimize its motions more globally, we propose to incorporate unconstrained transit motions between constrained rearranging actions. 
We extend our previous dhRRT \cite{ren2022rearrangement} to build a forest-based planning framework by concurrently planning local rearranging actions from different robot configurations. 
The proposed framework possesses progress-controlled planning horizons and reactivity to real-world uncertainties as the dhRRT does, but more importantly, is additionally able to:
\begin{enumerate}
    \item incorporate robot free motions between rearranging actions without being always constrained in $SE(2)$, enabling the robot to quickly switch between different problem subspaces to reduce unnecessary motions and improve the effectiveness of the generated actions;

    \item monitor the planning progress in different subspaces by concurrently spawning and growing multiple trees, so as to direct the planning into more task-relevant subspaces for improving the planning efficiency;
    
    \item utilize the gradients of task heuristics to quickly steer the planning towards the most task-relevant subspaces, to further facilitate the planning efficiency.
    
\end{enumerate}

\section{Related Work}
\label{sec:relatedwork}

\emph{Rearrangement Planning:} 
Many approaches to rearrangement planning use only pick-and-place actions\cite{alami1994two, simeon2004manipulation, stilman2007manipulation} to move each object without needing to consider the physics, solving the problem under simplified settings. 
By using physics-based planning, approaches were proposed to generate more diverse 
solutions to the problem.
For example, by using high-level motion primitives\cite{barry2013manipulation, dogar2012planning, mericli2013achievable}, rearrangement can be achieved by moving one object at a time.
By allowing concurrent multiple object-object interactions, \cite{king2015nonprehensile, haustein2015kinodynamic} developed Kinodynamic RRT-based algorithms to efficiently search in the high-dimensional problem space.
More recently, learning-based rearrangement planning has been extensively studied.
By learning a control policy from data, the robot is able to plan its motions online while observing the current state to achieve various manipulation tasks by rearrangement, such as pushing-based relocation\cite{yuan2019end}, multi-object rearrangement and singulation \cite{eitel2020learning, hermans12}, and rearrangement-based grasping \cite{laskey2016robot}.
However, such approaches in general require much data and time to train policies and are difficult to generalize to different tasks.

\emph{Planning under Uncertainties:}
Due to real-world physics and perception uncertainties, the real execution on the robot often moves the objects differently from what is predicted by any open-loop planner.
One solution to such problems is by integrating the uncertainty models into the planning process\cite{koval2015robust, johnson2016convergent, moll2017randomized}. 
Such solutions improve the probability of the execution success while being conservative on selecting motions.
Another solution is by interleaving the planning and execution to generate closed-loop robot motions with receding planning horizons\cite{bejjani2018planning}.
Similar to our previous work\cite{ren2022rearrangement}, in this work we also use heuristically controlled dynamic horizons to address real-world uncertainties.

\emph{Forest-based Planners:}
As being able to more flexibly and efficiently explore the problem space compared to tree-based planners, forest-based planners have been studied and deployed for different purposes and task requirements.
Early work constructs distributed roadmaps of trees, as a form of forest-based planner, to address large-scale motion planning problems\cite{plaku2005distributed, plaku2005sampling, akinc2005probabilistic}.
\cite{gayle2007lazy} maintains a forest of RRTs for being reactive to dynamical environment changes.
\cite{vonasek2019space} grows multiple trees to concurrently explore different configuration subspaces for efficient multi-goal path planning.
\cite{lai2021adaptively} adaptively creates new trees in constrained regions by exploiting the local structure learned from past observations, improving the sampling efficiency in highly constrained problem space with theoretical completeness and optimality guarantees.
However, these existing forest-based planners were initially designed for geometric path planning problems, without being able to easily adapt to a kinodynamic system.

\section{Problem Statement}
\label{sec:problem}

For rearrangement-based manipulation problems, we consider an environment to be rearranged to a state to satisfy certain task goal criterion.
The problem contains a robot manipulator, $N$ movable objects to rearrange, and a set of static obstacles denoted as $\mathcal{O}$. Moreover, we define two different types of robot motions:
\begin{enumerate}
    \item Rearranging actions: Such actions are constrained inside a task manifold, such that the end-effector moves in $SE(2)$ parallel to the support surface, to let the robot locally manipulate the surrounding movable objects.
    
    \item Transit motions: These are unconstrained robot motions in the robot's collision-free configuration space $Q^R_{free}$, as will be defined in Sec.~\ref{sec:terminology}. By such motions the robot moves without contacting anything in the workspace in $SE(3)$.
\end{enumerate}

In this work, 
different from the formulation in\cite{ren2022rearrangement},
we reformulate the rearrangement-based manipulation planning as a novel coupled local rearrangement planning and global action optimization problem, to more globally explore the problem space and optimize the search in a more directed way. 
As such, the solution to the problem is represented by a sequence of motion segments containing only local rearranging actions connected by segments of transit motions, as will be formally described in Sec.~\ref{sec:probdef}.
The transit motions between local rearranging actions can be solved efficiently by a geometric path planar such as RRT-connect\cite{kuffner2000rrt}.
One major challenge in such a formulation is to determine when and how to switch between different types of robot motions, which is important to the efficiency of the problem solving and the optimality of the solutions.


\subsection{Terminology}
\label{sec:terminology}

We solve the planning of local rearranging actions, termed as local rearrangement planning, by kinodynamic planning since sophisticated system dynamics has to be taken into account due to concurrent object-object interactions resulted from rearranging actions.
For this, we define:

\subsubsection{State Space}

The state space of the local rearrangement planning, defined as $\mathcal{Q}$, is composed of the configuration space of the robot and the state spaces of all the movable objects. 
Formally, the robot's configuration space is defined as $\mathcal{Q}^R \subset \mathbb{R}^r$, where $r \in \mathbb{Z}^+$ is the robot's degree of freedom.
Furthermore, we define the collision-free configuration space of the robot $\mathcal{Q}^R_{free} \subset \mathcal{Q}^R$ as the subspace in which the robot does not collide with itself or anything within the workspace.
The state space of the $i$-th movable object is $\mathcal{Q}^i \subset SE(2)$ where $i \in \{1, 2, ..., N\}$.
Therefore, the state space of the local rearrangement planning problem becomes the Cartesian product $\mathcal{Q} = \mathcal{Q}^R \times \mathcal{Q}^1 \times ... \times \mathcal{Q}^N$. 
And a state is represented by the tuple $q = (q^R, q^1, ..., q^N) \in \mathcal{Q}$, where $q^R \in \mathcal{Q}^R$ is the state of the robot and $q^i \in \mathcal{Q}^i$ is the state of the $i$-th movable object.
We define the valid state space $\mathcal{Q}^{valid} \subseteq \mathcal{Q}$ to include all the valid system states.
And for a state to be valid, all dimensions of the state must be within their bounds and there is no collision between the robot and itself or any static obstacle. 
It is worth noting that the contacts between movable objects and static obstacles and between any pair of movable objects or the robot are allowed.

\subsubsection{Control Space and Constraint Manifold} 

In the local rearrangement planning, we consider a control space $\mathcal{U} \subset \mathbb{R}^r$ consisting of all controls the robot is allowed to perform under some motion constraints.
Such motion constraints form a task manifold $\mathcal{M} \subset \mathcal{Q}$.
In our case, the robot's rearranging actions are constrained such that the end-effector moves planarly in $SE(2)$, resulting in a constraint task manifold where at each state the robot's end-effector remains the same height and zero roll and pitch.

\subsubsection{Transition Function}
The system physics is represented by a transition function, $\Gamma: \mathcal{Q}^{valid} \times \mathcal{U} \mapsto \mathcal{Q}$, which maps a state $q_t \in \mathcal{Q}^{valid}$ and a control $u_t \in \mathcal{U}$ at time $t$ to the outcome state at the next step $q_{t+1} \in \mathcal{Q}$.

\subsubsection{Goal Criterion} 
For any manipulation task, we generally represent the task goal by a criterion function $g: \mathcal{Q}^{valid} \mapsto \{0, 1\}$.
And the goal region of the planning problem is the subspace in which all states satisfy the goal criterion, denoted by $\mathcal{Q}_G = \{q\; | \;  g(q) = 1, \; q \in \mathcal{Q}^{valid} \}$.

\subsection{Local Rearrangement and Global Action Optimization}
\label{sec:probdef}


Formally, starting from the current system state $q_s \in \mathcal{Q}^{valid}$ at which the corresponding robot state is $q^R_s \in \mathcal{Q}^R_{free}$, the planning problem is resolved by finding a sequence of $M$ paired motion segments $\xi= \{(\bar{\tau}_1, \tau_1), ..., (\bar{\tau}_m, \tau_m), ..., (\bar{\tau}_M, \tau_M)\}$, where each pair consists of 
a transit motion segment $\bar{\tau}_m$ followed by a rearranging action segment $\tau_m$.
As aforementioned, a strategy of determining how to switch between these two types of motion segments is critical, which in general needs global action exploration and optimization.
The transit motion segment, as defined by $\bar{\tau}_m: [0, 1] \mapsto \mathcal{Q}^R_{free}$, is a geometric path plan in the robot's configuration space; and the rearranging action segment is represented by a sequence of $T$ system states and robot controls $\tau_m = \{q^m_0, u^m_0, ..., q^m_t, u^m_t, ..., q^m_T, u^m_T\}$ in which the system transits by the physics laws $\Gamma(q^m_t, u^m_t) = q^m_{t+1}$ for all $t = 0, \ldots, T$.
And the following conditions must be satisfied for a feasible solution plan:
\begin{itemize}
    \item The start state of the first motion segment is the same as the current robot state: $\bar{\tau}(0) = q^R_s$;
    \item The end state of the final motion segment satisfies the goal criterion: $\Gamma(q^M_T, u^M_T) \in \mathcal{Q}_G$;
    \item All the intermediate states along any transit motion segment are valid, i.e., $\forall m, \forall t \in [0, 1]: \bar{\tau}_m(t) \in \mathcal{Q}^R_{free}$;
    \item All the intermediate states along any rearranging action segment are valid and inside the constraint task manifold, i.e., $\forall m, \; \forall t: \; q^m_t \in \mathcal{Q}^{valid} \cap \mathcal{M}$;
    \item The end and start states of adjacent segments are the same, i.e., $\forall m: \; \bar{\tau}_m(1) = q^m_0$ and $q^{m}_T = \bar{\tau}_{m+1}(0)$.
\end{itemize}

Compared to the existing kinodynamic planning used in \cite{ren2022rearrangement}, the formulation above is more general for representing a rearrangement-based manipulation problem.
To see that, by forcing the entire motion plan to only contain rearranging action segments, i.e.,  $\forall m: \; \bar{\tau}_m = \{\}$, the problem will reduce to the traditional formulation; and as a result, the robot will always move inside the constraint manifold $\mathcal{M}$ and the end-effector will continuously translate and rotate in $SE(2)$ without being able to quickly reposition itself to explore different task-relevant subspaces.

\section{Kinodynamic Rapidly-exploring Random Forest}
\label{sec:method}

\begin{figure}[t]
    \centering
        \includegraphics[width=\columnwidth]{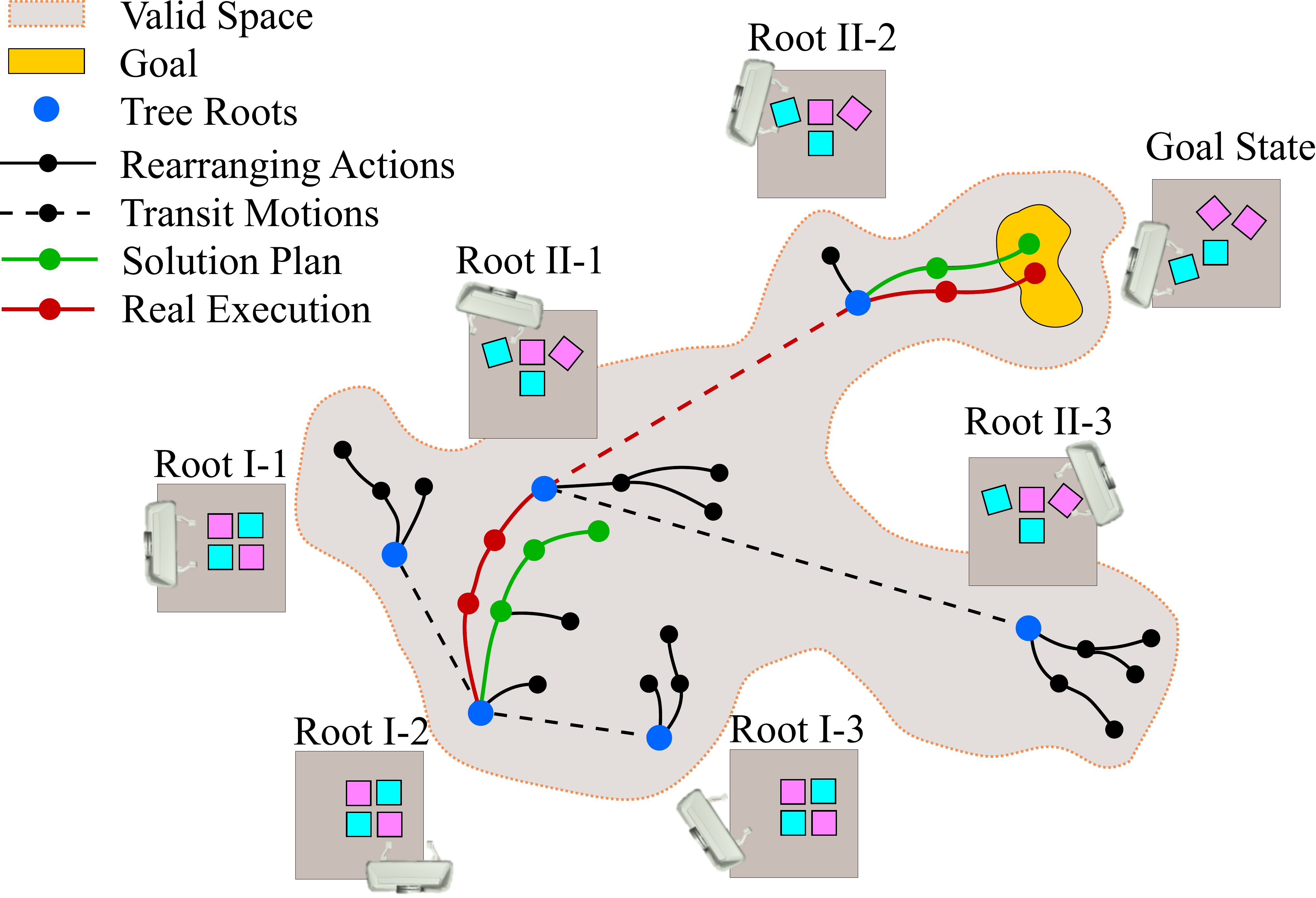}
    \caption{A schematic plot of kdRRF. By spawning and growing multiple trees rooted with different robot configurations, the kdRRF more efficiently explores task-relevant subspaces and selects motions (green) from a more global perspective.
    At the same time, by interleaving planning and real execution (red), the errors resulted from uncertainties are eliminated in real time.}
    \label{fig:kdRRF}
    \vspace{-0.5cm}
\end{figure}

To address the problem formulated in Sec.~\ref{sec:problem}, our framework leverages the sampling-based kinodynamic RRT (kdRRT) planner \cite{lavalle2001randomized} for local rearrangement planning, to take advantage of its efficiency in high-dimensional state spaces and generalizability to various kinodynamic planning problems.
However, instead of growing one individual motion tree, our framework concurrently grows multiple trees to form a forest $F$.
By that, our framework is able to concurrently explore different subspaces starting from different robot configurations and more globally optimize the generated motions.
As such, we name our framework as Kinodynamic Rapidly-exploring Random Forest, abbreviated as $kdRRF$.
The algorithm is summarized in Alg.~\ref{alg:kdrrf} and illustrated in Fig.~\ref{fig:kdRRF}.
And the details will be described below.

\begin{figure}[t]
\vspace{-0.2cm}
\begin{algorithm}[H]
\caption{The kdRRF algorithm}
\footnotesize
    \begin{algorithmic}[1]
        \Require Start state $q_s$, goal region $\mathcal{Q}_G(\cdot)$, heuristic $h(\cdot)$, progress threshold $p$, forest size limit $S_{max}$, number of trees $N_{tree}$
        \Ensure Motion segments $\xi$
        \State $F \gets \Call{SpawnForest}{N_{tree}, q_s}$
        \Comment{Alg.~\ref{alg:spawnforest}}
        \State $\xi \gets \{\}$, $q^* \gets q_s$
        \While{$\Call{Time.Available}{\null}$}
            \State $F \gets \Call{ExpandForest}{F}$
            \Comment{Alg.~\ref{alg:expandtree}}
            \State $(\bar{\tau}, \tau) \gets \Call{EvaluateProgress}{F, q*, h, p, S_{max}}$
            \Comment{Alg.~\ref{alg:evaluateprogress}}
            \If{$\tau \neq \{\}$}
                \State $\Call{FollowRobotPath}{\bar{\tau}}$
                \Comment{In $\mathcal{Q}^R_{free}$}
                \State $ q^* \gets \Call{ExecuteControls}{\tau}$
                \Comment{Observe Real State}
                \State $\xi \gets \xi \cup (\bar{\tau}, \tau)$
                \If{$q^* \in \mathcal{Q}_G$}
                \Comment{Task Complete}
                \State \textbf{break}
                \EndIf
                \State $F \gets \Call{SpawnForest}{N_{tree}, q^*}$
                \Comment{Alg.~\ref{alg:spawnforest}}
            \EndIf
        \EndWhile
        \State \Return $\xi$
    \end{algorithmic}
    \label{alg:kdrrf}
\end{algorithm}
\vspace{-1cm}
\end{figure}

\begin{figure}[t]
\vspace{-0.2cm}
\begin{algorithm}[H]
\caption{SpawnForest($\cdot$)}
\footnotesize
    \begin{algorithmic}[1]
        \Require Number of trees $N_{tree}$, state $q$
        \Ensure Initialized forest $F$
        \State $F.\Call{AddRoot}{q}$
        \State $q^R \gets \Call{ExtractRobotState}{q}$
        \For{$i = 2, ..., N_{tree}$}
            \While{True}
                \State $q_i \gets \Call{SampleRoot}{q}$
                \Comment{Sec.~\ref{sec:rootsample}}
                \State $q^R_i \gets \Call{ExtractRobotState}{q_i}$
                \If{$q^R_i \in \mathcal{Q}^R_{free}$ and $\Call{ExistsPath}{q^R, q^R_i}$}
                    \State $F.\Call{AddRoot}{q_i}$
                    \State \textbf{break}
                \EndIf
            \EndWhile
        \EndFor
        \State \Return $F$
    \end{algorithmic}
    \label{alg:spawnforest}
\end{algorithm}
\vspace{-1cm}
\end{figure}

In each tree of the forest $F$, every node represents a system state and every edge represents a control $u_t$.
As illustrated in Alg.~\ref{alg:spawnforest}, given the start state, a number of $N_{tree}$ tree roots will be spawned by randomly sampling valid end-effector's poses in the workspace, as detailed in Sec.~\ref{sec:rootsample}.
And then the forest will be expanded by adding new nodes into one of its trees, as detailed in Alg.~\ref{alg:expandtree}. 
Given any randomly sampled state $q_{rand}$, the nearest node in the forest, $q_{near}$, will be selected to expand. 
From $q_{near}$, a set of $C$ controls will be randomly sampled, denoted as $\{u^1, \ldots, u^C\}$. 
And one control $u^*$ from this set, which minimizes the distance between $\Gamma(q_{near}, u^*)$ and $q_{rand}$, will be selected to grow the forest. 
We simultaneously monitor the status of each tree in the forest while expanding, 
and will temporarily stop the forest expansion when some criteria have been met, as detailed in Sec.~\ref{sec:kdrrf} and illustrated in Alg.~\ref{alg:evaluateprogress}.
Then one tree in the forest will be selected based on the progress made by this tree to extract control segment for execution.
This entire process of interleaving planning and execution will be repeated until the task goal is achieved.

We use a heuristic function: $h: \mathcal{Q}^{valid} \mapsto \mathbb{R}$ to monitor the system progress in the planning process.
When the state of the system moves closer to the task goal, $h(\cdot)$ will decrease.
In practice, we do not require the heuristics to be optimal.
Simple cost-decreasing functions can be easily integrated in our planning framework to successfully drive the search to find solutions, as will be described in Sec.~\ref{sec:app}.



\begin{figure}[t]
\begin{algorithm}[H]
\caption{ExpandForest($\cdot$)}
\footnotesize
    \begin{algorithmic}[1]
        \Require Current forest $F$
        \Ensure Expanded forest $F$
        \State $q_{rand} \gets \Call{SampleState()}{}$
        \State $q_{near} \gets \Call{FindNearest}{F, q_{rand}}$
        \For{$i = 1, ..., C$}
            \State $v_i \gets \Call{SampleControl()}{}$
            \Comment{In $se(2)$}
            \State $q_i \gets \Gamma(q_{near}, v_i)$
            \Comment{State Transition}
        \EndFor
        \State $(q_{new}, v^*) \gets \mathrm{\arg\min}_{(q_i, v_i)} \Call{Distance}{q_i, q_{rand}}$
        \State $u^* \gets \Call{JocobianProjection}{v^*}$
        \Comment{Sec.~\ref{sec:jocobian}}
        \If{$u^* \neq Null$}
            \State $F.\Call{AddNode}{q_{new}}$
            \Comment{Expansion}
            \State $F.\Call{AddEdge}{(q_{near}, q_{new}), u^*}$
        \EndIf
        \State \Return $F$
    \end{algorithmic}
    \label{alg:expandtree}
\end{algorithm}
\vspace{-1cm}
\end{figure}

\begin{figure}[t]
\vspace{-0.2cm}
\begin{algorithm}[H]
\caption{EvaluateProgress($\cdot$)}
\footnotesize
    \begin{algorithmic}[1]
        \Require Current forest $F$, current system state $q_t$, heuristic $h(\cdot)$, progress threshold $p$, forest size limit $S_{max}$
        \Ensure Paired motion segment $(\bar{\tau}, \tau)$
        \State $q_{new} \gets F.\Call{GetLatestNode}{\null}$
        \State $\bar{\tau} \gets \{\}$, $\tau \gets \{\}$
        \If{$q_{new} \in \mathcal{Q}_G$ \textbf{or} $h(q_t) - h(q_{new}) > p$}
            \State $q_{root} \gets F.\Call{TraceTreeRoot}{q_{new}}$
            \State $\bar{\tau} \gets \Call{GeneratePath}{q_t, q_{root}}$
            \Comment{In $\mathcal{Q}^R_{free}$}
            \State $\tau \gets \Call{ExtractControls}{F, q_{new}}$
        \ElsIf{$F.\Call{GetSize()}{} = S_{\max}$}
        \Comment{Forest Size Limit}
            \State $q^\prime \gets \mathrm{\arg\min}_{q \in F.\Call{GetLeaves}{\null}} h(q)$
            \State $q_{root} \gets F.\Call{TraceTreeRoot}{q^\prime}$
            \State $\bar{\tau} \gets \Call{GeneratePath}{q_t, q_{root}}$
            \Comment{In $\mathcal{Q}^R_{free}$}
            \State $\tau \gets \Call{ExtractControls}{F, q^\prime}$
        \EndIf
        \State \Return $(\bar{\tau}, \tau)$ 
    \end{algorithmic}
    \label{alg:evaluateprogress}
\end{algorithm}
\vspace{-1cm}
\end{figure}

\subsection{Sampling Tree Roots in the Forest}
\label{sec:rootsample}

At the beginning of the planning process, the forest is constructed by sampling multiple roots to spawn multiple motion trees.
One of the tree roots always corresponds to the current system state, whereas the other tree roots are generated by randomly sampling different valid end-effector's poses in $SE(2)$ with the states of all the movable objects unchanged.
The robot's configuration of each tree root, given the sampled pose of the end-effector, is solved by inverse kinematics of the robot. 
A tree root is considered valid only when there exists some path in $\mathcal{Q}^R_{free}$ connecting the robot's current configuration to the configuration of this tree root, which can be efficiently checked by a geometric path planner like RRT-connect\cite{kuffner2000rrt}.

\emph{Task-oriented Root Sampling.} As making contacts with movable objects is essential for exploring effective rearranging actions, the end-effector's pose is sampled near the movable objects.
Specifically, for sampling a new tree root, we first randomly select a movable object $i \in \{1, \ldots, N\}$, and place the end-effector near the selected object to some ranged distance.
The selection of the movable object can be guided by a discrete uniform distribution by default, i.e., $P(i) = 1 / N$, where $P(i)$ denotes the probability of selecting the $i$-th movable object. 
Alternatively, the movable object can be sampled by the following probability mass function depending on the gradient of the heuristic function with respect to each object state $q^i$, which can be analytically derived or numerically approximated with small local changes on $q^i$. We term this strategy as Task-oriented Root Sampling, as it uses the task heuristics to effectively bias the local search.
\begin{equation}
\begin{aligned}
    P(i) = \frac{f\left(\lvert \nabla_{q^i} h \rvert\right)}{Z}, \quad i \in \{1, \ldots, N\},
\end{aligned}
\end{equation}
where $f: \mathbb{R}^+ \mapsto \mathbb{R}^+$ is a stretching function and $Z = \sum_{i} f\left(\lvert \nabla_{q^i} h \rvert\right)$ is the normalization term. 
In practice, we use $f(\cdot)$ to introduce some desired properties of the distribution for sampling. 
For example, we use exponential, i.e., $f(x) = e^x$, to avoid zero probability for sampling any object; or we use power functions, i.e., $f(x) = x^k$, to stretch when the gradients are non-zero for all the objects.


\subsection{Dynamic Planning Horizons}
\label{sec:kdrrf}

Due to physical and perception uncertainties in the local rearrangement planning, the real-world execution can deviate from the planned path to cause execution failures.
To address this problem, the kdRRF interleaves the planning and execution to close the manipulation loop similarly to our previous work \cite{ren2022rearrangement}. 
The planning horizon is dynamically adjusted by the planning progress evaluated by the heuristics $h(\cdot)$.


While we grow the forest by expanding the nearest tree, we monitor the heuristic value of the newly added node.
Once good enough progress is made by this new node, determined by a threshold $p \in \mathbb{R}$, we will back-trace this node to extract the paired motion segment from the tree and execute it on the robot, as detailed in Alg.~\ref{alg:evaluateprogress}.
Otherwise, when it is difficult to make enough progress starting from the current state, we will limit the planning process by another threshold, $S_{max}$, of the maximum forest size -- the total number of nodes in the forest.
In this case, the leaf node with the best heuristic score in the forest will be selected for execution, if not enough progress can be made but $S_{max}$ has been reached.
In this manner, the horizon for local rearrangement planning is dynamically determined by the heuristics $h(\cdot)$ in terms of the current system state and tree selection.

For each execution, the robot will first transit itself to the configuration associated with the root node of the selected tree by transit motions, and then subsequently execute the extracted control segment by constrained rearrange actions.
After each execution, our framework will observe the actual system state from sensor readings, and then repeat the same procedure to progressively transit the system towards the goal region.
Rather than planning and executing the entire motion plan at one time, our framework intermittently updates the planner with the most recent system state to continuously eliminate the errors in the real-world executions in Alg.~\ref{alg:kdrrf}.

\subsection{Constrained Projection via Jacobian}
\label{sec:jocobian}


To impose the motion constraints for rearranging actions, we sample robot controls only in the end-effector's velocity space $v \in se(2)$, and iteratively project it to the robot's control space $\mathcal{U}$ via Jacobian by $u = J^\dag(q^R) \cdot v$, as the robot state $q^R$ changes throughout performing the control. 

\section{Example Applications}
\label{sec:app}


\begin{figure}[t]
\footnotesize
\centering
    \begin{tikzpicture}
        \node[anchor=south west,inner sep=0] at (0,0){\includegraphics[width=\linewidth]{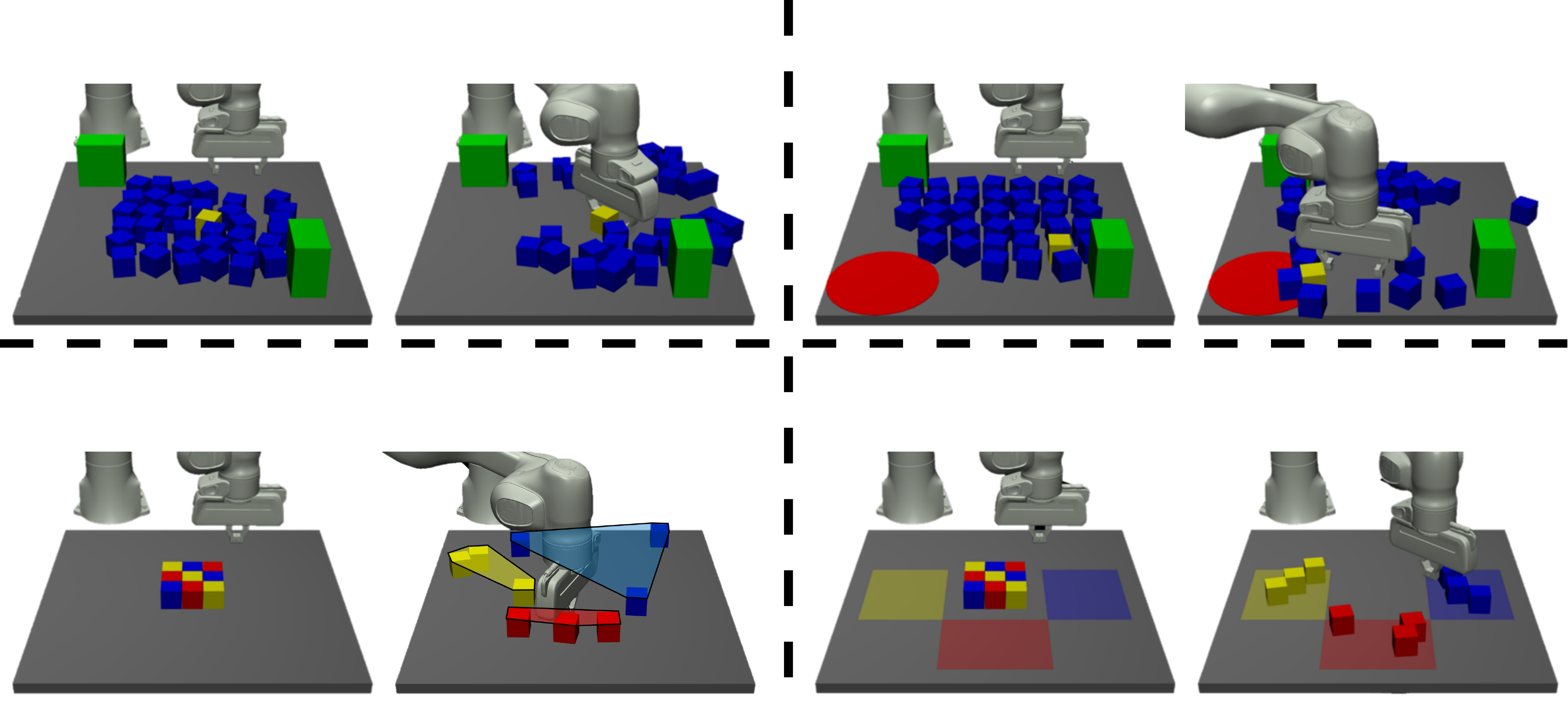}};
        \node[align=center] at (2.1, 3.7) {Grasping};
        \node[align=center] at (6.5, 3.7) {Relocating};
        \node[align=center] at (2.1, 1.7) {Sorting without Goal Regions};
        \node[align=center] at (6.5, 1.7) {Sorting with Goal Regions};
    \end{tikzpicture}
    \caption{The $4$ example rearrangement-based planning tasks defined in Sec.~\ref{sec:app}. The left figure shows the start state, and the right figure shows the goal state achieved by our proposed planner. For Grasping and Relocating: Green (obstacles), Blue (movable objects), Yellow (target object).}
\label{fig:app}
\vspace{-0.5cm}
\end{figure}

Our framework is evaluated with $4$ rearrangement-based manipulation tasks in clutter, as exemplified in Fig.~\ref{fig:app}: 1) \emph{Grasping}, 2) \emph{Relocating}, 3) \emph{Sorting without Goal Regions}, and 4) \emph{Sorting with Goal Regions}.
The first three tasks and their heuristic functions are defined the same as in our previous work\cite{ren2022rearrangement}.
Additionally, we add the fourth task as defined below:

\emph{Sorting with Goal Regions:}
This sorting task
specifies a goal region for each class of the objects.
We denote by $L$ the number of object classes, and by $B_i = \{j \in \{1, \ldots, N\}: \; \Call{Class}{j} = i \}$ the set of all objects in the $i$-th class. 
The goal region $\mathcal{G}_i$ for the $i$-th class is a rectangle centered at $(x_{\mathcal{G}_i}, y_{\mathcal{G}_i})$.
The goal for this sorting task is achieved when all the movable objects are relocated to be inside their corresponding goal region.
Formally, $\forall i \in \{1, \ldots, L\}$:
\begin{equation}
\begin{aligned}
   \forall j \in \mathcal{B}_i, \; (x^j, y^j) \in \mathcal{G}_i
\end{aligned}
\end{equation}
where $(x^j, y^j)$ is the position of the $j$-th object in $SE(2)$.

The heuristic function for this task is defined by the summation of the squared distance between any object and the center of its corresponding goal region:
\begin{equation}
\begin{aligned}
    h_s(q) = \sum_{1 \leq i\leq L} \sum_{j \in \mathcal{B}_i} (x^j - x_{\mathcal{G}_i})^2 + (y^j - y_{\mathcal{G}_i})^2
\end{aligned}
\end{equation}
\section{Experiments}
\label{sec:experiments}

\begin{figure*}[t]
\setlength{\tabcolsep}{5pt}
\centering
\footnotesize
\centering
\begin{tabular}{c|| c c | c c | c c | c c}
    \hline
    \multirow{2}{*}{} & \multicolumn{2}{c|}{Grasping} & \multicolumn{2}{c|}{Relocating} & \multicolumn{2}{c|}{Sorting without Goal Regions} & \multicolumn{2}{c}{Sorting with Goal Regions}\\
    \cline{2-9}
    & dhRRT & kdRRF & dhRRT & kdRRF & dhRRT & kdRRF & dhRRT & kdRRF\\
    \hline
    Success Rate & \: 84 / 100 & \textbf{\: 96 / 100} & \: 88 / 100 & \textbf{\: 95 / 100} & \: 62 / 100 & \textbf{\: 99 / 100} & \: 19 / 100 & \textbf{ 100 / 100}\\
    Time (seconds) & 43.1 $\pm$ 39.9 & \textbf{20.2 $\pm$ 18.9} & \: 32.3 $\pm$ 39.0 & \textbf{23.5 $\pm$ 19.4} & 111.4 $\pm$ \: 76.0 & \textbf{\: 31.1 $\pm$ 14.3} & 215.3 $\pm$ 52.4 & \textbf{\: 48.9 $\pm$ 10.9}\\
    Num. Action & 40.1 $\pm$  33.4 & \textbf{11.2 $\pm$ \: 9.7} & \: 37.1 $\pm$ 36.5 & \textbf{16.1 $\pm$ \: 9.5} & 227.7 $\pm$ 108.2 & \textbf{\: 54.9 $\pm$ 14.8} & 311.5 $\pm$  76.0 & \textbf{\: 54.9 $\pm$ \: 8.5}\\
    \hline
\end{tabular}
\vspace{1pt}
\caption{Experiment results of the $4$ tasks in simulation.}
\label{tab:sim_eval0}
\vspace{-0.3cm}
\end{figure*}

\begin{figure*}[t]
\setlength{\tabcolsep}{5pt}
\centering
\footnotesize
\centering
\begin{tabular}{c|| c c | c c | c c | c c}
    \hline
    \multirow{2}{*}{} & \multicolumn{2}{c|}{Grasping} & \multicolumn{2}{c|}{Relocating} & \multicolumn{2}{c|}{Sorting without Goal Regions} & \multicolumn{2}{c}{Sorting with Goal Regions}\\
    \cline{2-9}
    & dhRRT & kdRRF & dhRRT & kdRRF & dhRRT & kdRRF & dhRRT & kdRRF\\
    \hline
    Success Rate & \: 7 / 10 & \textbf{\: 9 / 10} & \textbf{10 / 10} & \: 7 / 10 & \: 1 / 10 & \textbf{\: 7 / 10} & \: 0 / 10 & \textbf{\: 9 / 10}\\
    Time (seconds) & \textbf{14.8 $\pm$ \: 3.5} & 25.8 $\pm$ 13.4 & \textbf{\: 25.3 $\pm$ 11.5} & 29.4 $\pm$ 14.5 & \: 54.8 $\pm$ \: 0.0 \:  & \textbf{30.7 $\pm$ 10.4} &  \: \: \: \: -- \: \: \: \:  & \textbf{103.8 $\pm$ 49.0}\\
    Num. Action & 83.9 $\pm$ 20.9 & \textbf{40.6 $\pm$ 19.9} & 105.8 $\pm$ 37.2 & \textbf{49.6 $\pm$ 11.5} & \: 63.0 $\pm$ \: 0.0 \: & \textbf{34.3 $\pm$ 11.2} &  \: \: \: \: -- \: \: \: \:  & \textbf{\: 46.1 $\pm$ 11.2}\\
    \hline
\end{tabular}
\vspace{1pt}
\caption{Experiment results of the $4$ tasks in the real world.}
\label{tab:real_eval}
\vspace{-0.5cm}
\end{figure*}

We evaluate our framework with the $4$ tasks defined in Sec.~\ref{sec:app} from two aspects: 1) the planning efficiency, reflected by the success rate given a limited planning time budget and average planning time for successful trials; and 2) the effectiveness of the generated motions, 
for which we report the average number of rearranging actions
for successful motion plans.
We expect a smaller number of 
rearranging actions for more efficient motion plans generated by the planner, as it requires less effort from the robot for rearranging objects to achieve the task goal.
The experiments were conducted both in the MuJoCo simulator\cite{todorov2012mujoco} and on a real Franka Panda Emika robot.
In both experiments, we used MuJoCo physics engine for transition function computation. 

In addition, we compared our new algorithm with a baseline dhRRT planner\cite{ren2022rearrangement}, 
which has shown better performance on planning efficiency and robustness against real-world uncertainties than traditional kinodynamic planners.
As described in Sec.~\ref{sec:rootsample}, 
we evaluated our algorithm with Task-oriented Root Sampling to demonstrate the effectiveness of the design for this technique.
All the algorithms were implemented in Python and run with single thread on a 3.4 GHz AMD Ryzen 9 5950X CPU.
For control sampling in all the experiments, 
the linear velocity of the end-effector was sampled within $[-0.2, 0.2] m/s$ in simulation, but $[-0.1, 0.1] m/s$ on the real robot for better safety; and the angular velocity was all sampled within $[-1, 1] rad/s$.

\vspace{-0.1cm}

\subsection{Simulation Evaluations}

In simulation, we used $N = 36$ objects for grasping and sorting, 
one of which was the target object.
For both sorting tasks, we used $L=3$ classes and $3$ objects in each class, 
resulting in $N=9$ objects to rearrange.
The planning time budget was set to $180$ seconds for grasping and relocating, and $300$ seconds for both sorting tasks.
For each task, we ran $100$ trials to collect
the statistics of relevant metrics.

The results are reported in Fig.~\ref{tab:sim_eval0}.
We can see that the kdRRF implementation achieved better success rate and planning time than dhRRT in all the evaluated tasks.
For the two sorting tasks, dhRRT has only $62 \%$ and $19 \%$ success rates respectively while kdRRF almost succeeds all the time.
Moreover, benefited from the global optimization of actions, kdRRF can generate 
more effective rearranging actions.
For grasping and relocating, kdRRF averagely requires less than half the number of rearranging actions as dhRRT; and for both sorting tasks, dhRRT has to plan more than $200$ number of rearranging actions to achieve the same goal while kdRRF only needs less than $60$ rearranging actions.
Importantly, as we observed when conducting the experiments, the Task-oriented Root Sampling guided the planner to explore more extensively in the most task-relevant subspaces, increasing the probability of finding effective motion segments.

\subsection{Real-world Evaluations}

\begin{figure}[t]
    \centering
    \footnotesize
    \begin{tikzpicture}
        \node[anchor=south west,inner sep=0] at (0,0){\includegraphics[width=\columnwidth]{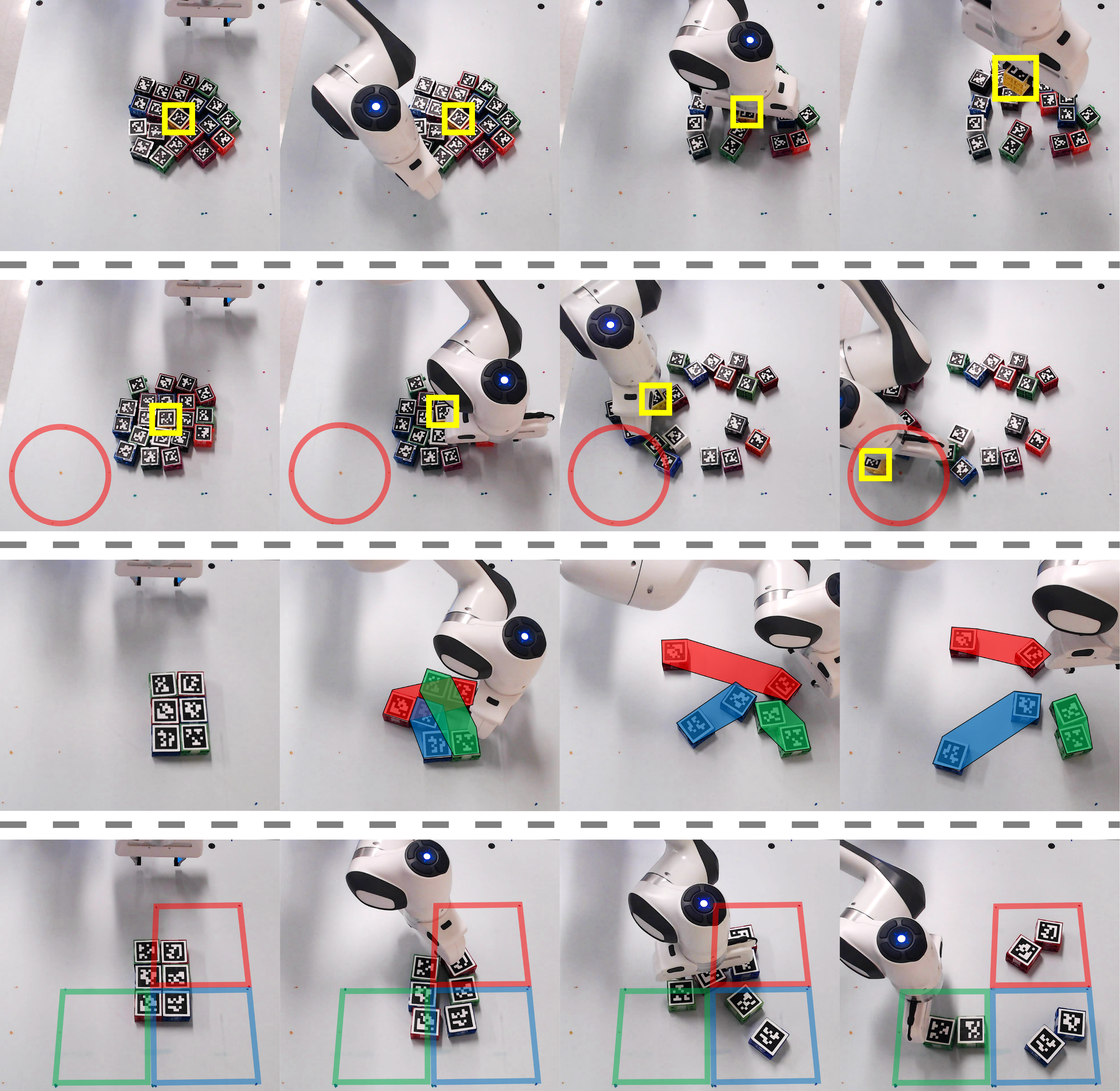}};
        \node[anchor=west, align=left] at (0, 8.2) {Grasping};
        \node[anchor=west, align=left] at (0, 6) {Relocating};
        \node[anchor=west, align=left] at (0, 3.6) {Sorting \\without \\Goals};
        \node[anchor=west, align=left] at (0, 1.4) {Sorting\\ with \\Goals};
    \end{tikzpicture}
    \caption{Real-world experiments on 4 rearrangement-based manipulation tasks. The target object in grasping and relocating tasks is highlighted by the yellow bounding box, and the convex hulls in the sorting task are represented by the overlays in different colors. }
    \label{fig:realexps}
\vspace{-0.5cm}
\end{figure}

We conducted experiments on a real robot to more realistically challenge the algorithms by real-world physical and perception uncertainties.
We used cube-like objects for rearrangement.
However, the physical parameters, including the size and the mass of the object, the friction coefficient of the table surface, etc, were very roughly estimated.
We used $N=20$ objects for grasping and relocating, and $L = 3$ classes and $N=6$ objects for both sorting tasks, two objects in each class.
The planning time budget was set to $60$ seconds for grasping, relocating, and sorting without goal regions, and to $180$ seconds for sorting with goal regions.
All the objects were tracked via AprilTags\cite{olson11}.
Example executions for the $4$ tasks on the real robot are shown in Fig.~\ref{fig:realexps}.
We ran $10$ trials for each evaluation with random initial states, and summarized the results in Fig.~\ref{tab:real_eval}.

We note that, for grasping and relocating which are relatively simpler tasks than sorting, kdRRF did not improve the planning efficiency against dhRRT as the simulation results show. This is because, under complex real-world uncertainties, kdRRF needed a lot more planning cycles than it did in simulation to adaptively correct accumulated physical errors to finish the task, resulting in longer total planning time. Nevertheless, we can still see that the number of rearranging actions needed by kdRRF is significantly less than dhRRT, indicating more effective executions.
For both sorting tasks, being consistent with the experiments in simulation, kdRRF outperformed the dhRRT from all aspects.
With a lot more rearranging actions and contacts with the objects, the manipulation by dhRRT often ran out of the time budget and barely succeeded. In contrast, kdRRF could more efficiently find most effective actions and achieved much higher success rates.

\section{Conclusion}
\label{sec:conclusion}

In this work, we formulated the rearrangement-based manipulation problem as a coupled problem of local rearrangement and global action optimization by incorporating free-space transit motions between constrained rearranging actions of the robot. 
For efficiently solving the rearrangement-based manipulation problem formulated as such, we proposed a forest-based kinodynamic planning framework that globally explores multiple problem subspaces to optimize its planning, and generates actions by concurrently growing multiple motions trees, while interleaving the planning and execution to react to uncertainties. 

We evaluated the proposed framework with extensive experiments on $4$ manipulation tasks both in simulation and on a real robot, the proposed kdRRF framework has shown significant improvement on planning efficiency and motion effectiveness compared to a state-of-the-art baseline, without sacrificing the robustness against real-world uncertainties.

In future work, we plan to further extend this framework to take into account other state features besides heuristic scores, such as the number of concurrent contacts made by a tree edge, to improve the quality of the generated motions and the planner's robustness against uncertainties. 
Moreover, as using a forest provides more feasibility for multi-threading, we also consider parallelizing the implementation of the algorithm to speed up its computation.

\bibliographystyle{IEEEtran}
\bibliography{refs}

\begin{thebibliography}{10}
\providecommand{\url}[1]{#1}
\csname url@samestyle\endcsname
\providecommand{\newblock}{\relax}
\providecommand{\bibinfo}[2]{#2}
\providecommand{\BIBentrySTDinterwordspacing}{\spaceskip=0pt\relax}
\providecommand{\BIBentryALTinterwordstretchfactor}{4}
\providecommand{\BIBentryALTinterwordspacing}{\spaceskip=\fontdimen2\font plus
\BIBentryALTinterwordstretchfactor\fontdimen3\font minus
  \fontdimen4\font\relax}
\providecommand{\BIBforeignlanguage}[2]{{%
\expandafter\ifx\csname l@#1\endcsname\relax
\typeout{** WARNING: IEEEtran.bst: No hyphenation pattern has been}%
\typeout{** loaded for the language `#1'. Using the pattern for}%
\typeout{** the default language instead.}%
\else
\language=\csname l@#1\endcsname
\fi
#2}}
\providecommand{\BIBdecl}{\relax}
\BIBdecl

\bibitem{agboh2018real}
W.~C. Agboh and M.~R. Dogar, ``Real-time online re-planning for grasping under
  clutter and uncertainty,'' in \emph{IEEE International Conference on Humanoid
  Robots (HUMANOIDS)}.\hskip 1em plus 0.5em minus 0.4em\relax IEEE, 2018, pp.
  1--8.

\bibitem{huang2019large}
E.~Huang, Z.~Jia, and M.~T. Mason, ``Large-scale multi-object rearrangement,''
  in \emph{2019 International Conference on Robotics and Automation
  (ICRA)}.\hskip 1em plus 0.5em minus 0.4em\relax IEEE, 2019, pp. 211--218.

\bibitem{moll2017randomized}
M.~Moll, L.~Kavraki, J.~Rosell \emph{et~al.}, ``Randomized physics-based motion
  planning for grasping in cluttered and uncertain environments,'' \emph{IEEE
  Robotics and Automation Letters}, vol.~3, no.~2, pp. 712--719, 2017.

\bibitem{ren2022rearrangement}
K.~Ren, L.~E. Kavraki, and K.~Hang, ``Rearrangement-based manipulation via
  kinodynamic planning and dynamic planning horizons,'' in \emph{IEEE
  International Conference on Intelligent Robots and Systems (IROS)}, 2022.

\bibitem{king2015nonprehensile}
J.~E. King, J.~A. Haustein, S.~S. Srinivasa, and T.~Asfour, ``Nonprehensile
  whole arm rearrangement planning on physics manifolds,'' in \emph{IEEE
  International Conference on Robotics and Automation (ICRA)}.\hskip 1em plus
  0.5em minus 0.4em\relax IEEE, 2015, pp. 2508--2515.

\bibitem{haustein2015kinodynamic}
J.~A. Haustein, J.~King, S.~S. Srinivasa, and T.~Asfour, ``Kinodynamic
  randomized rearrangement planning via dynamic transitions between statically
  stable states,'' in \emph{IEEE International Conference on Robotics and
  Automation (ICRA)}.\hskip 1em plus 0.5em minus 0.4em\relax IEEE, 2015, pp.
  3075--3082.

\bibitem{king2017unobservable}
J.~E. King, V.~Ranganeni, and S.~S. Srinivasa, ``Unobservable monte carlo
  planning for nonprehensile rearrangement tasks,'' in \emph{IEEE International
  Conference on Robotics and Automation (ICRA)}.\hskip 1em plus 0.5em minus
  0.4em\relax IEEE, 2017, pp. 4681--4688.

\bibitem{alami1994two}
R.~Alami, J.-P. Laumond, and T.~Sim{\'e}on, ``Two manipulation planning
  algorithms,'' in \emph{WAFR Proceedings of the workshop on Algorithmic
  foundations of robotics}.\hskip 1em plus 0.5em minus 0.4em\relax AK Peters,
  Ltd. Natick, MA, USA, 1994, pp. 109--125.

\bibitem{simeon2004manipulation}
T.~Sim{\'e}on, J.-P. Laumond, J.~Cort{\'e}s, and A.~Sahbani, ``Manipulation
  planning with probabilistic roadmaps,'' \emph{The International Journal of
  Robotics Research}, vol.~23, no. 7-8, pp. 729--746, 2004.

\bibitem{stilman2007manipulation}
M.~Stilman, J.-U. Schamburek, J.~Kuffner, and T.~Asfour, ``Manipulation
  planning among movable obstacles,'' in \emph{Proceedings 2007 IEEE
  international conference on robotics and automation}.\hskip 1em plus 0.5em
  minus 0.4em\relax IEEE, 2007, pp. 3327--3332.

\bibitem{barry2013manipulation}
J.~Barry, K.~Hsiao, L.~P. Kaelbling, and T.~Lozano-P{\'e}rez, ``Manipulation
  with multiple action types,'' in \emph{Experimental Robotics}.\hskip 1em plus
  0.5em minus 0.4em\relax Springer, 2013, pp. 531--545.

\bibitem{dogar2012planning}
M.~R. Dogar and S.~S. Srinivasa, ``A planning framework for non-prehensile
  manipulation under clutter and uncertainty,'' \emph{Autonomous Robots},
  vol.~33, no.~3, pp. 217--236, 2012.

\bibitem{mericli2013achievable}
T.~A. Mericli, M.~M. Veloso, and H.~L. Akin, ``Achievable push-manipulation for
  complex passive mobile objects using past experience.'' in \emph{AAMAS},
  2013, pp. 71--78.

\bibitem{yuan2019end}
W.~Yuan, K.~Hang, D.~Kragic, M.~Y. Wang, and J.~A. Stork, ``End-to-end
  nonprehensile rearrangement with deep reinforcement learning and
  simulation-to-reality transfer,'' \emph{Robotics and Autonomous Systems},
  vol. 119, pp. 119--134, 2019.

\bibitem{eitel2020learning}
A.~Eitel, N.~Hauff, and W.~Burgard, ``Learning to singulate objects using a
  push proposal network,'' in \emph{Robotics research}.\hskip 1em plus 0.5em
  minus 0.4em\relax Springer, 2020, pp. 405--419.

\bibitem{hermans12}
T.~Hermans, J.~M. Rehg, and A.~Bobick, ``Guided pushing for object
  singulation,'' in \emph{IEEE International Conference on Intelligent Robots
  and Systems (IROS)}, 2012, pp. 4783--4790.

\bibitem{laskey2016robot}
M.~Laskey, J.~Lee, C.~Chuck, D.~Gealy, W.~Hsieh, F.~T. Pokorny, A.~D. Dragan,
  and K.~Goldberg, ``Robot grasping in clutter: Using a hierarchy of
  supervisors for learning from demonstrations,'' in \emph{IEEE international
  conference on automation science and engineering}.\hskip 1em plus 0.5em minus
  0.4em\relax IEEE, 2016, pp. 827--834.

\bibitem{koval2015robust}
M.~C. Koval, J.~E. King, N.~S. Pollard, and S.~S. Srinivasa, ``Robust
  trajectory selection for rearrangement planning as a multi-armed bandit
  problem,'' in \emph{2015 IEEE/RSJ International Conference on Intelligent
  Robots and Systems (IROS)}.\hskip 1em plus 0.5em minus 0.4em\relax IEEE,
  2015, pp. 2678--2685.

\bibitem{johnson2016convergent}
A.~M. Johnson, J.~E. King, and S.~Srinivasa, ``Convergent planning,''
  \emph{IEEE Robotics and Automation Letters}, vol.~1, no.~2, pp. 1044--1051,
  2016.

\bibitem{bejjani2018planning}
W.~Bejjani, R.~Papallas, M.~Leonetti, and M.~R. Dogar, ``Planning with a
  receding horizon for manipulation in clutter using a learned value
  function,'' in \emph{2018 IEEE-RAS 18th International Conference on Humanoid
  Robots (Humanoids)}.\hskip 1em plus 0.5em minus 0.4em\relax IEEE, 2018, pp.
  1--9.

\bibitem{plaku2005distributed}
E.~Plaku and L.~E. Kavraki, ``Distributed sampling-based roadmap of trees for
  large-scale motion planning,'' in \emph{Proceedings of the 2005 IEEE
  International Conference on Robotics and Automation}.\hskip 1em plus 0.5em
  minus 0.4em\relax IEEE, 2005, pp. 3868--3873.

\bibitem{plaku2005sampling}
E.~Plaku, K.~E. Bekris, B.~Y. Chen, A.~M. Ladd, and L.~E. Kavraki,
  ``Sampling-based roadmap of trees for parallel motion planning,'' \emph{IEEE
  Transactions on Robotics}, vol.~21, no.~4, pp. 597--608, 2005.

\bibitem{akinc2005probabilistic}
M.~Akinc, K.~E. Bekris, B.~Y. Chen, A.~M. Ladd, E.~Plaku, and L.~E. Kavraki,
  ``Probabilistic roadmaps of trees for parallel computation of multiple query
  roadmaps,'' in \emph{Robotics Research. The Eleventh International
  Symposium}.\hskip 1em plus 0.5em minus 0.4em\relax Springer, 2005, pp.
  80--89.

\bibitem{gayle2007lazy}
R.~Gayle, K.~R. Klingler, and P.~G. Xavier, ``Lazy reconfiguration forest
  (lrf)-an approach for motion planning with multiple tasks in dynamic
  environments,'' in \emph{Proceedings 2007 IEEE International Conference on
  Robotics and Automation}.\hskip 1em plus 0.5em minus 0.4em\relax IEEE, 2007,
  pp. 1316--1323.

\bibitem{vonasek2019space}
V.~Von{\'a}sek and R.~P{\v{e}}ni{\v{c}}ka, ``Space-filling forest for
  multi-goal path planning,'' in \emph{2019 24th IEEE International Conference
  on Emerging Technologies and Factory Automation (ETFA)}.\hskip 1em plus 0.5em
  minus 0.4em\relax IEEE, 2019, pp. 1587--1590.

\bibitem{lai2021adaptively}
T.~Lai and F.~Ramos, ``Adaptively exploits local structure with generalised
  multi-trees motion planning,'' \emph{IEEE Robotics and Automation Letters},
  vol.~7, no.~2, pp. 1111--1117, 2021.

\bibitem{kuffner2000rrt}
J.~J. Kuffner and S.~M. LaValle, ``Rrt-connect: An efficient approach to
  single-query path planning,'' in \emph{Proceedings 2000 ICRA. Millennium
  Conference. IEEE International Conference on Robotics and Automation.
  Symposia Proceedings (Cat. No. 00CH37065)}, vol.~2.\hskip 1em plus 0.5em
  minus 0.4em\relax IEEE, 2000, pp. 995--1001.

\bibitem{lavalle2001randomized}
S.~M. LaValle and J.~J. Kuffner~Jr, ``Randomized kinodynamic planning,''
  \emph{The international journal of robotics research}, vol.~20, no.~5, pp.
  378--400, 2001.

\bibitem{todorov2012mujoco}
E.~Todorov, T.~Erez, and Y.~Tassa, ``Mujoco: A physics engine for model-based
  control,'' in \emph{IEEE International Conference on Intelligent Robots and
  Systems (IROS)}.\hskip 1em plus 0.5em minus 0.4em\relax IEEE, 2012, pp.
  5026--5033.

\bibitem{olson11}
E.~Olson, ``Apriltag: A robust and flexible visual fiducial system,'' in
  \emph{IEEE International Conference on Robotics and Automation (ICRA)}, 2011,
  pp. 3400--3407.

\end{thebibliography}

\end{document}